%% file: main.tex
\documentclass[conference]{IEEEtran}
\IEEEoverridecommandlockouts

\usepackage{siunitx} 
\usepackage[utf8]{inputenc} 
\usepackage[T1]{fontenc} 
\usepackage{amssymb} 
\usepackage{graphicx} 
\usepackage[nolist]{acronym} 
\usepackage[backend = biber, defernumbers = false, style = ieee, citestyle = ieee, dashed = false]{biblatex} 
\usepackage[babel]{csquotes} 
\usepackage{hyperref} 
\usepackage{booktabs} 
\usepackage{pgfplots} 
\usepackage{xstring}
\usepackage{nomencl}
\makenomenclature
\usepackage{amsmath,amssymb,amsfonts}
\usepackage{algorithmic}
\usepackage{graphicx}
\usepackage{textcomp}
\usepackage{xcolor}
\def\BibTeX{{\rm B\kern-.05em{\sc i\kern-.025em b}\kern-.08em
    T\kern-.1667em\lower.7ex\hbox{E}\kern-.125emX}}

\include{Settings/Configuration.tex}

\include{Settings/Custom_commands.tex}

\include{Kapitel/Acronyms/Acronyms.tex}

\begin{document}
    \title{ARIDA - Active railway infrastructure dataset annotation using neural network predictions\\}
        \thanks{This research has received funding from the Federal Ministry for Economic Affairs and Climate Action (BMWK) under grant agreements 19I21039A.}

    \author{\IEEEauthorblockN{Maximilian~Dornik$^{1}$, Christian~Sieberichs$^{2}$, Simon~Geerkens$^{2}$ and Alexander~Braun$^{2}$\\[0.35ex]
        \small
        $^{1}${\texttt{maximilian.dornik@study.hs-duesseldorf.de}}\\
        $^{2}${\texttt{\{christian.sieberichs, simon.geerkens, alexander.braun\}@hs-duesseldorf.de}}} University of Applied Sciences Düsseldorf\\ Düsseldorf, Germany}

    \maketitle
    \thispagestyle{empty}
    \pagestyle{empty}

    \begin{abstract}
        Modern railway perception requires large comprehensive image datasets, but creating them is slowed by manual labeling, especially for track polylines and other complex infrastructure geometry.
        This paper presents the Active Railway Infrastructure Dataset Annotation framework, a railway-specific semi-automated pipeline that converts instance predictions from a mask scoring instance segmentation network into bounding-boxes, polygons, polylines, and paired polylines.
        To keep automatic track geometry annotation reliable without ground truth, it applies confidence thresholds and an unique contour-agreement quality score as an acceptance gate.
        An active learning loop ranks remaining images by uncertainty and adapts the number of manual queries per cycle.
        This is achieved through a novel quantity control mechanism, in which easy images are auto-accepted and human effort targets hard cases.
        Experiments on disjoint, diverse imagery and time-sequential, redundant imagery quantify effort reduction and accuracy trade-offs, achieving about two thirds reduction in the disjoint regime and up to \perc{93} in the sequential regime.
        The study also provides guidance for initial label budgets and stopping to avoid over-annotation.
    \end{abstract}


    \input{Kapitel/Introduction/Introduction.tex}

    \input{Kapitel/Related_Work/Related_Work.tex}

    \input{Kapitel/Materials_and_Methods/Materials_and_Methods.tex}

    \input{Kapitel/Results/Results.tex}

    \input{Kapitel/Discussion/Discussion.tex}

    \input{Kapitel/Conclusion/Conclusion.tex}

    \printbibliography

\end{document}

%% file: Settings/Configuration.tex
        \DeclareSourcemap{
            \maps[datatype=bibtex]{
                \map[overwrite=false]{
                    \step[fieldset=year,fieldvalue={n.y.},final] 
                }
            }
        }

    \newlength{\imgWidth} 

%% file: Settings/Custom_commands.tex
    \newcommand{\Hypothese}[1]{\hyperref[Hypothese #1]{\textbf{H#1}}}

        \newcommand{\mean}{\mathop{\mathrm{mean}}}

        \newcommand{\std}{\mathop{\mathrm{std}}}

        \newcommand{\perc}[1]{#1\,\%}

%% file: Kapitel/Acronyms/Acronyms.tex
\begin{acronym}[arida]
    \acro{al}[AL]{Active Learning}
    \acro{map}[mAP]{Mean Average Precision}
    \acro{arida}[ARIDA]{Active Railway Infrastructure Dataset Annotation}
    \acro{tbal}[TBAL]{Triplets based Active Learning}
    \acro{msrcnn}[MS R-CNN]{Mask Scoring region-based convolutional neural network}
    \acro{cfs}[CFS]{Contour F Score}
    \acro{roi}[ROI]{Region of Interest}
    \acro{iou}[IoU]{Intersection over Union}
    \acro{ml}[ML]{Machine Learning}
    \acro{cv}[CV]{Computer Vision}
    \acro{ai}[AI]{Artificial Intelligence}
    \acro{osdar23}[OSDaR23]{Open Sensor Data for Rail 2023}
    \acro{dzsf}[DZSF]{German Centre for Rail Traffic Research at the Federal Railway Authority}
    \acro{cvat}[CVAT]{Computer Vision Annotation Tool}
    \acro{via}[VIA]{VGG Image Annotator}
    \acro{mit}[MIT]{Massachusetts Institute of Technology}
    \acro{asam}[ASAM]{Association for Standardization of Automation and Measuring Systems}
    \acro{yolop}[YOLOP]{You Only Look Once for Panoptic Driving Perception}
    \acro{sam}[SAM]{Segment Anything Model}
    \acro{lidar}[LiDAR]{Light Detection and Ranging}
    \acro{roi}[RoI]{Regions of Interest}
    \acro{rpn}[RPN]{Region Proposal Network}
    \acro{iou}[IoU]{Intersection over Union}
    \acro{qqp}[QQP]{Query Quantity Parameterization}
    \acro{adas}[ADAS]{Advanced Driver-Assistance Systems}
    \acro{mrcnn}[Mask R-CNN]{Mask region-based convolutional neural network}
\end{acronym}

%% file: Kapitel/Introduction/Introduction.tex
\section{Introduction}
    \label{section: Introduction}

    Computational perception through \ac{ai} is a safety-critical enabler for \ac{adas} and higher levels of automation in railway transportation.
    Reliable scene understanding supports functions such as obstacle detection, track availability assessment, and operational decision-making, where errors can have direct safety and economic consequences. \cite{peleska2022standardisation}\cite{zeller_safetrain_2023}\cite{vdv2019operation}
    While deep learning has substantially advanced visual perception in other domains, progress in autonomous railway perception remains comparatively limited, and one of the primary bottlenecks is the availability of large, high-quality annotated datasets \cite{zendel2019railsem19}\cite{tagiew2023osdar23}.

    A central challenge is, that public railway datasets are not only scarce \cite{pappaterra2021datasets}, but the available datasets, most notably \ac{osdar23} \cite{tagiew2023osdar23} and the currently most comprehensive RailSem19 \cite{zendel2019railsem19}, are often insufficient for training modern computer vision systems at scale.
    Such existing public datasets are limited in size and do not provide the data variety or annotation primitives required to learn robust models for real-world operation.
    At the same time, creating improved or more advanced railway datasets in an own endeavour is challenging with current tooling. \cite{hiebert2025labels4rails}\cite{tagiew2023osdar23}

\begin{figure}[!h]
        \begin{centering}
            \begin{minipage}[t]{0.1175\textwidth}
                \centering
                    \includegraphics[width = 0.125\imgWidth, keepaspectratio = true]{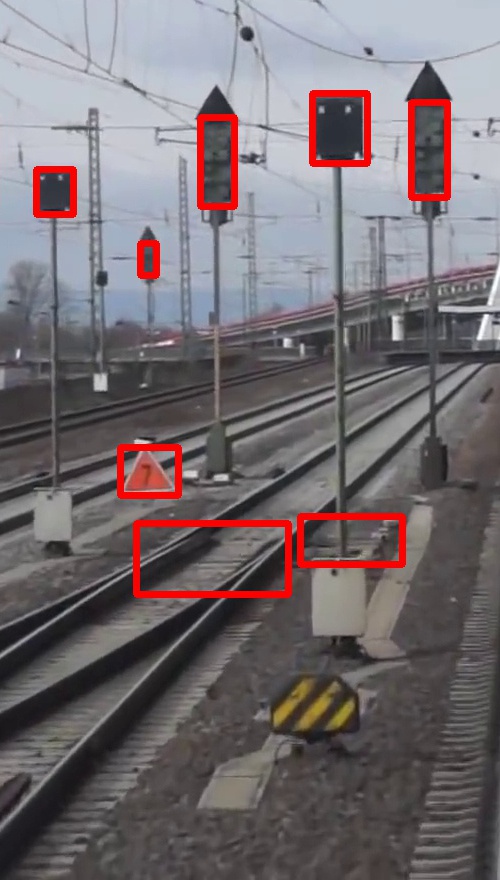}
                    Bounding-Box
            \end{minipage}
            \begin{minipage}[t]{0.1175\textwidth}
                \centering
                    \includegraphics[width = 0.125\imgWidth, keepaspectratio = true]{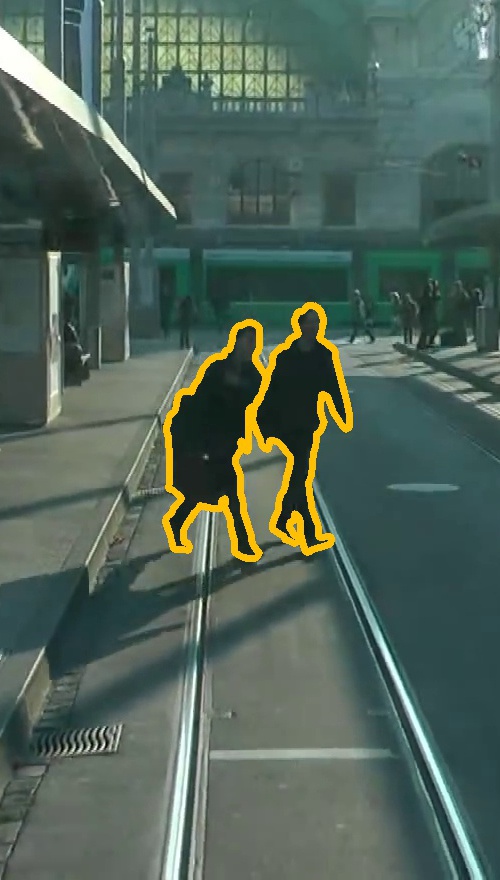}
                    Polygon
            \end{minipage}
            \begin{minipage}[t]{0.1175\textwidth}
                \centering
                    \includegraphics[width = 0.125\imgWidth, keepaspectratio = true]{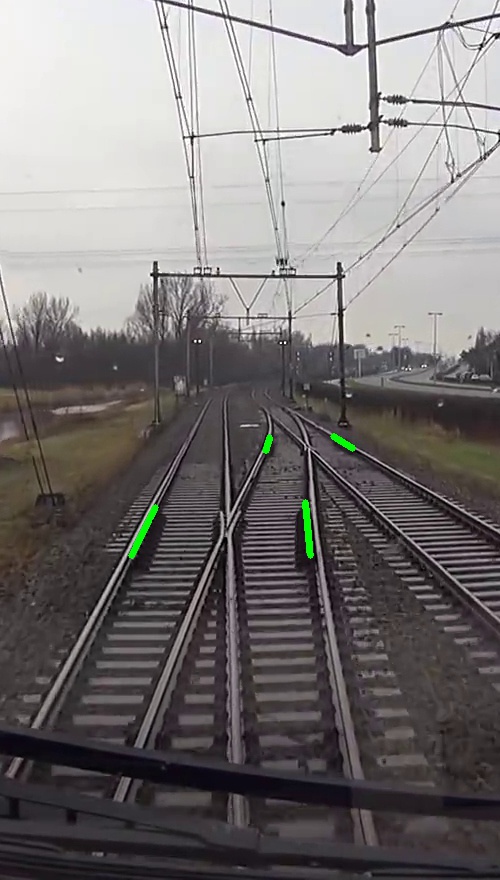}
                    Polyline
            \end{minipage}
            \begin{minipage}[t]{0.1175\textwidth}
                \centering
                    \includegraphics[width = 0.125\imgWidth, keepaspectratio = true]{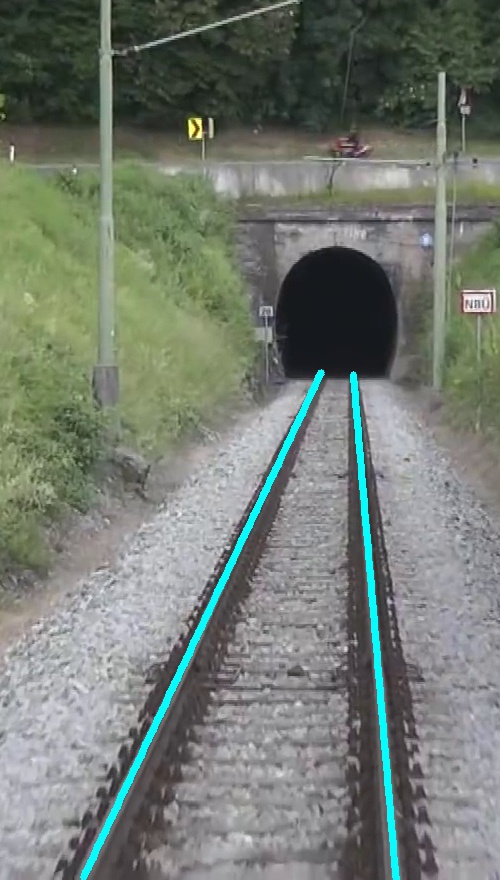}
                    Polyline-Pair
            \end{minipage}
        \end{centering}
        \caption[Cropped example images of the annotation types contained in the RailSem19 dataset]{Cropped example images of the annotation types contained in the RailSem19 dataset. Only a single annotation type with an arbitrarily chosen color is shown per image for demonstration purposes. Images source: \cite{zendel2019railsem19}}
        \label{Grafik: AnnotationTypesRailSem19}
    \end{figure}

    Railway perception commonly requires instance-based geometric annotations beyond standard bounding-boxes or masks (s. Figure \ref{Grafik: AnnotationTypesRailSem19}), including coordinate-based polylines and polyline-pairs that represent track geometry \cite{zendel2019railsem19}.

    These primitives are essential for track understanding yet are particularly labor-intensive to produce \cite{hiebert2025labels4rails}\cite{zendel2019railsem19}.
    Moreover, many general-purpose annotation tools like LabelMe \cite{noauthor_labelme_nodate} or the \ac{cvat} \cite{noauthor_open_nodate} (s. related work in section \ref{section: Related Work}) are not tailored to railway requirements and do not support the full set of required primitives.
    In addition, when annotation budgets are constrained, it is not sufficient to merely assist labeling.
    One must also prioritize which samples should be labeled to maximize performance gains per human annotation.
    This creates a reinforcing limitation: datasets remain suboptimal for training, but producing better datasets is prohibitively expensive and technically cumbersome. \cite{hiebert2025labels4rails}\cite{settles2009active}\cite{li2025deepactivelearning}
    These constraints motivate the following research question \textbf{RQ}:

    \textbf{RQ:} How can manual annotation effort be significantly reduced when creating high-quality railway image datasets with multi-primitive annotations suitable for training modern perception models?

    In this work, we use annotation effort to denote the human workload required to transform raw railway imagery into a usable training set. This includes the frequency with which human annotators must intervene to correct or complete labels, and the overall share of images that require manual labeling rather than simple verification.
    Reducing annotation effort therefore means reducing the amount of manual geometric drawing and limiting human intervention to genuinely difficult cases, while still producing accurate, application-relevant annotations.


    To address the \textbf{RQ}, we propose the \ac{arida} framework.
    \ac{arida} is a semi-automated image annotation framework that combines standalone model-based annotation capabilities with data-efficient sample selection.
    \ac{arida} is centered around an instance-segmentation model (\ac{msrcnn} \cite{huang_mask_2019}), that generates per-instance predictions.
    An annotation module then converts these predictions into fully featured RailSem-compatible primitives with a dedicated procedure to derive polyline-pair track geometries from predicted track masks.
    We choose RailSem-style primitives because they represent the most complete publicly established railway annotation scheme and include the track-critical polyline-pair primitives that many pipelines do not support \cite{zendel2019railsem19}\cite{hiebert2025labels4rails}.
    To ensure that automatically generated track annotations are reliable without requiring ground truth, \ac{arida} includes a no-ground-truth quality gate based on our \ac{cfs}.
    Finally, \ac{arida} integrates an \ac{al} loop that prioritizes informative and difficult samples for human labeling, so that the annotator effort is concentrated where it yields the highest benefit \cite{settles2009active}\cite{li2025deepactivelearning}.
    To our knowledge, no prior railway-specific framework combines semi-automatic multi-primitive annotation generation, track-polyline-pair quality gating without ground truth, and an integrated \ac{al}-loop.

    We hypothesize firstly, that \ac{arida} reduces manual annotation effort compared to fully manual labeling by automatically producing a substantial fraction of annotations and routing only low-confidence or low-quality cases to human annotators.
    Secondly, we hypothesize, that \ac{al} improves sample efficiency for railway instance segmentation, achieving comparable model performance with fewer human-labeled images than non-selective supervised training \cite{settles2009active}.

    The main contributions of this paper are:

    \begin{enumerate}
        \item \textbf{\ac{arida} framework:} A railway-specific semi-automatic pipeline with focus on automated annotation designed to reduce manual dataset-creation effort with an instance segmentation and iterative \ac{al} core.
        \item \textbf{Multi-primitive annotation module:} Conversion of predictions into RailSem-style primitives, with a dedicated focus on polyline-pair track annotation.
        \item \textbf{\ac{cfs} quality control:} A no-ground-truth metric used as an acceptance gate for automatically generated track polyline-pairs.
        \item \textbf{\ac{al} evaluation baseline:} First time evaluation of \ac{al} for railway instance segmentation, providing a baseline for future research.
        \item \textbf{Empirical guidance:} Guidance for parameterization of \ac{arida} on commonly occuring disjoint versus time-sequential datasets in the railway domain.
    \end{enumerate}

    The remainder of this paper is organized as follows: Section \ref{section: Related Work} reviews related work, Section \ref{section: Materials and Methods} presents \ac{arida} and its components, while Section \ref{section: Results} describes the experimental setup and evaluation results. In Section \ref{section: Discussion}, findings and limitations are discussed with Section \ref{section: Conclusion} concluding the paper.

%% file: Kapitel/Related_Work/Related_Work.tex
\section{Related Work}
    \label{section: Related Work}

    The railway domain continues to face a pronounced data bottleneck.
    Public datasets for railway perception are comparatively rare and often limited in size, scope, or annotation richness.
    Creating new datasets remains costly because high-quality railway annotations are labor-intensive to produce. \cite{pappaterra2021datasets}\cite{hiebert2025labels4rails}
    Prior work addresses this problem from two complementary directions.
    Upstream approaches aim to reduce human effort during annotation, whereas downstream approaches aim to reduce the amount of labeled data needed to reach a target model performance.
    To keep the discussion directly comparable to \ac{arida}, we structure the related work along the following dimensions:

    \begin{itemize}
        \item \textbf{Domain specificity:} Is the method railway-focused or general-purpose?
        \item \textbf{Data modality:} Is the approach designed for image data and/or other data types (e.g. \ac{lidar})?
        \item \textbf{Annotation primitives:} Which primitives are supported, and are they instance-level applicable?
        \item \textbf{Automation level:} To what extent does the approach reduce manual annotation effort?
        \item \textbf{Sample selection:} Does the approach include a method to prioritize samples for annotation?
    \end{itemize}

    \subsection{Upstream optimization: Annotation procedures}
        General-purpose semi-automated annotation tools follow an assistive paradigm.
        Models provide suggestions that humans verify and correct, improving suggestions over time.
        \ac{cvat} exemplifies this approach and supports image and video annotation with suggestion workflows for bounding-boxes and polygons \cite{noauthor_open_nodate}.
        However, such tools are not tailored to railway-specific topology, semantics, or workflows, which limits their efficiency for railway image annotation despite their broad primitive support.

        The strongest recent railway-specific upstream baseline is Labels4Rails \cite{hiebert2025labels4rails}, a railway image annotation tool for track geometry, switches, and scene tags with automation for ego-track annotation and switch derivation.
        Compared with \ac{cvat}, it reports weighted-average per-image annotation time reductions of \perc{42} in manual mode and \perc{58} with automated annotation and manual amendment.
        However, it remains centered on semi-automated labeling under continuous human supervision and its reported automation is limited in scope. \cite{hiebert2025labels4rails}
        In contrast, \ac{arida} targets automatic generation of all railway primitives considered here and uses quality gating to restrict human effort to uncertain or low-quality cases.


    \subsection{Downstream optimization: Data-efficient \acl{al}}
        \ac{al} addresses the data bottleneck from the opposite direction.
        By selecting the most informative unlabeled samples in a dataset for human annotation, model improvement per labeled sample can be maximized. \cite{settles2009active}

        In the railway domain, Denzler et al. demonstrate this for multi-sensor, sequence-based object detection.
        Using uncertainty estimation to select low-confidence sequences leveraging temporal multi-sensor context and label interpolation, they report reaching fully-annotated-dataset performance with \perc{25} fewer annotations and outperforming a random baseline by 2 percentage points.
        While this validates \ac{al}'s downstream value, the approach is sequence-dependent, focused on bounding-box detection, and not designed as an image-dataset annotation pipeline supporting richer railway primitives. \cite{denzler_multi-sensor_2022}

        Wulff et al. \cite{wulff2024railwaylidar} illustrate the synergy of upstream bootstrapping and downstream \ac{al}, but in a \ac{lidar}-centric setting.
        They use model-generated labels and an \ac{al}-loop to focus manual effort on the most informative \ac{lidar} scans.
        110 of 602 scans required manual correction after five cycles achieving \perc{71.48} performance, yet their pipeline is not designed for image annotation while also not supporting an image-centered multi-primitive output. \cite{wulff2024railwaylidar}

    \subsection{Railway track detection procedures}
        Even with progress in assisted annotation and \ac{al}, fine-grained track geometry remains one of the most labor-intensive parts of railway dataset creation.
        Track detection models can in principle provide reusable geometry for annotation automation, but they differ substantially in how directly their outputs align with annotation needs.

        \subsubsection{\textbf{Segmentation approaches}} Segmentation approaches like RailNet (Wang et al. \cite{DBLP:journals/access/WangWHQ19}) output pixel-wise track-region masks \cite{DBLP:journals/access/WangWHQ19}.
        From \ac{arida}'s viewpoint, they are image-based and railway-specific, but their outputs are class masks rather than coordinate primitives.
        Producing polylines and polyline-pairs requires additional post-processing, whereby instance separation is not inherent.

        \subsubsection{\textbf{Hybrid approaches}} Hybrid approaches like the method of Tao et al. \cite{tao_accurate_2021} combine learned predictions with deterministic geometry extraction to yield coordinate-based outputs that are closer to annotation needs.
        Such approaches recover rail coordinates from intermediate representations and output polylines, sometimes with instance separation. \cite{tao_accurate_2021}
        However, these methods are typically dedicated perception pipelines, focusing on operational use (e.g. ego-drivepath extraction) and a single primitive type, rather than being designed as a multi-primitive annotation pipeline with sample selection.

        \subsubsection{\textbf{End-to-end deep learning approaches}} End-to-end approaches directly regress coordinate-based rail representations and can produce instance-separated polylines (and sometimes polyline-pairs), but they still typically output only a single primitive family and do not address dataset annotation holistically.
        DALNet (Yu et al. \cite{yu_dalnet_2023}) for instance, yields instance-separated polylines without post-processing via an anchor-line formulation \cite{yu_dalnet_2023}.
        In principle, these outputs could be converted into polyline-pairs and other primitives, making the representation compatible with annotation needs.

    Taken together, recent railway-specific upstream work has reduced the gap to generic tools, with Labels4Rails demonstrating substantial time savings for railway image annotation \cite{hiebert2025labels4rails}.
    Downstream studies further show that \ac{al} can lower labeling demand in railway perception \cite{denzler_multi-sensor_2022}\cite{wulff2024railwaylidar}.
    Track-detection methods provide increasingly annotation-compatible geometry, but remain perception pipelines rather than dataset-construction frameworks \cite{tao_accurate_2021}\cite{yu_dalnet_2023}.
    \ac{arida} addresses the remaining gap by combining railway-specific multi-primitive annotation, quality-gated automation, and model-driven sample selection.

%% file: Kapitel/Materials_and_Methods/Materials_and_Methods.tex
\section{Materials and Methods}
    \label{section: Materials and Methods}

    To address the data scarcity and tooling limitations identified in the Introduction and Related Work, we propose the \ac{arida} framework.
    \ac{arida} is a comprehensive semi-automated pipeline for annotating railway image datasets efficiently.
    This section describes the framework modules and their internal procedures in detail.
    For the automatic annotation capability, we focus on rail-track annotation as it is the most challenging task and remains insufficiently studied in this principle.

\begin{figure}[!h]
        \centering
        \includegraphics[width = 0.5\imgWidth, keepaspectratio = true]{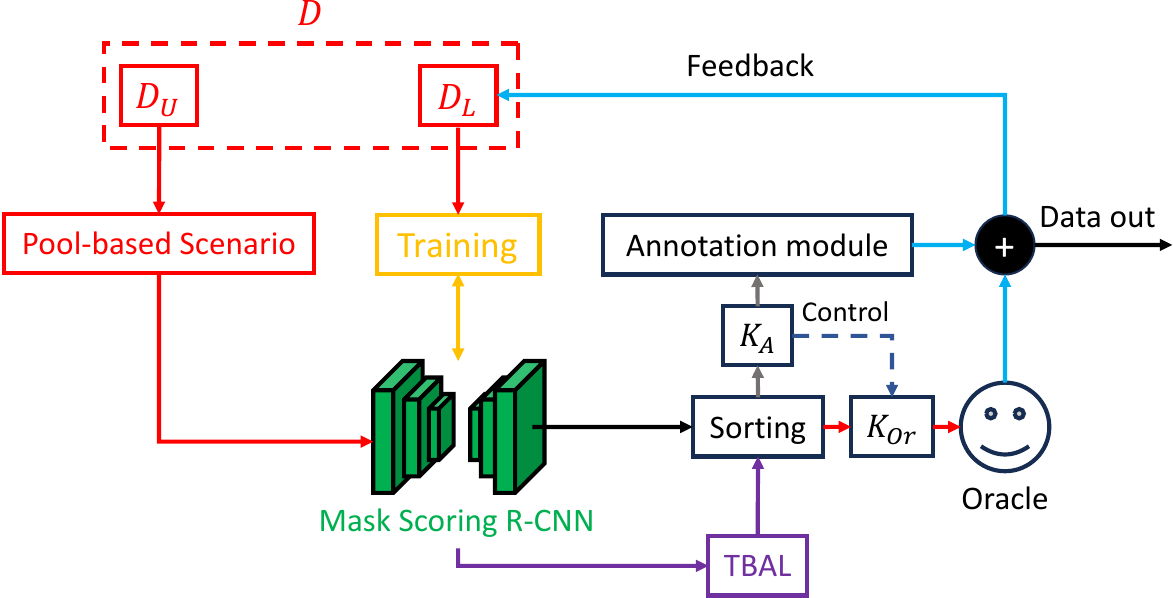}
        \caption[Simplified overview of the \ac{arida}-Framework architecture.]{Simplified overview of the \ac{arida}-framework architecture. All framework modules and data flows comprising \ac{al} are shown in red, while raw \ac{msrcnn} predictions are shown in black or grey (after sorting). Automatic annotations obtained by the annotation module and manual human annotations are shown in light-blue. Numerical procedure control data for and from the \acs{tbal}-module is shown in violet. The scalar control signal used for the Query Quantity Parameterization is shown in dark blue.}
        \label{Grafik: Framework Architecture}
    \end{figure}

    \ac{arida} implements a semi-automatic workflow that reduces manual labeling effort using an \ac{ai}-model.
    The model generates instance-level predictions that are converted into structured annotations.
    Figure \ref{Grafik: Framework Architecture} provides a simplified overview of the system architecture.

    The core of \ac{arida} combines \ac{msrcnn} with an annotation module.
    This base system enables automatic generation of instance-based annotations from model predictions.
    It produces bounding-boxes, polygons, polylines, and polyline-pairs, as illustrated in Figure \ref{Grafik: AnnotationTypesRailSem19}, using the hybrid approach.

    We adopt a hybrid design because most related approaches from the previous Section generate either semantic segmentation or a single annotation primitive.
    Producing all required primitives with separate specialized models would substantially increase training time, memory consumption, and system complexity.
    Such a multi-model setup would also require additional aggregation and post-processing for quality assurance and \ac{al}.
    Moreover, many existing approaches do not provide instance-based annotations, which our annotation policy requires.
    Our hybrid design therefore combines one instance-based model with deterministic conversion to deliver the full set of annotation types.

    \subsection{Mask Scoring R-CNN}
        \ac{msrcnn} \cite{huang_mask_2019} is a descendant of the \ac{mrcnn} \cite{he_mask_2018} network architecture principle designed for efficient instance segmentation of up to 100 image objects via a concatenated two-stage approach (object detection and instance segmentation) \cite{huang_mask_2019}\cite{he_mask_2018}.

        Firstly, \ac{roi}, proposed by the \ac{rpn}, are projected on image feature maps, which are then resized into a square shape through the \ac{roi}-align component.
        These square feature maps are then evaluated by a regression network to predict class scores and bounding-box coordinates for up to 100 object instances.
        The second network stage uses the object detection results to create semantic segmentation masks for each detected image object instance. \cite{he_mask_2018}
        \ac{msrcnn} adds a "mask\acs{iou}" network extension additionally, which regresses an \ac{iou} quality score of the predicted masks \cite{huang_mask_2019}, giving \ac{arida} a more reliable indicator on the mask predictions quality compared to the normally referred object classification score.
        This is especially important, as the mask quality directly influences the annotation quality in the subsequent annotation module.

        \ac{msrcnn} provides qualitative instance-level predictions, including bounding-boxes and segmentation masks, which our annotation module converts into the required railway primitives.
        Its dedicated mask\ac{iou} head explicitly estimates mask quality, enabling reliable filtering of low-quality predictions and supporting uncertainty and informativeness scoring for \ac{al}.
        \ac{msrcnn} is also available in established toolchains, can be fine-tuned with standard hyperparameters, and supports specialized \ac{al} strategies \cite{xu_active_2022}.
        Finally, it remains a strong-performing instance-segmentation baseline, making it a suitable backbone for semi-automated annotation.

    \subsection{Annotation module and the \acl{cfs}}
        The annotation module is part of the base system and converts raw \ac{msrcnn} predictions to the RailSem19 data format using four annotation type specific coordinate generation algorithms.
        Because railway tracks are among the most important objects to annotate \cite{zendel2019railsem19}, we focus mainly on the polyline-pair annotation type generation algorithm in this section.
        Other primitives are generated using substeps of this algorithm (polygons through contour coordinates, polylines through the centerline) or directly from the model predictions (bounding-boxes).
        The scheme for annotating railway tracks (label: "rail") through the polyline-pair annotation algorithm is exemplarily shown for a single object instance in the following Figure \ref{Grafik: Polyline-pair annotation pipeline}.

        \begin{figure}[!h]
            \centering
            \includegraphics[width = 0.49\imgWidth, keepaspectratio = true]{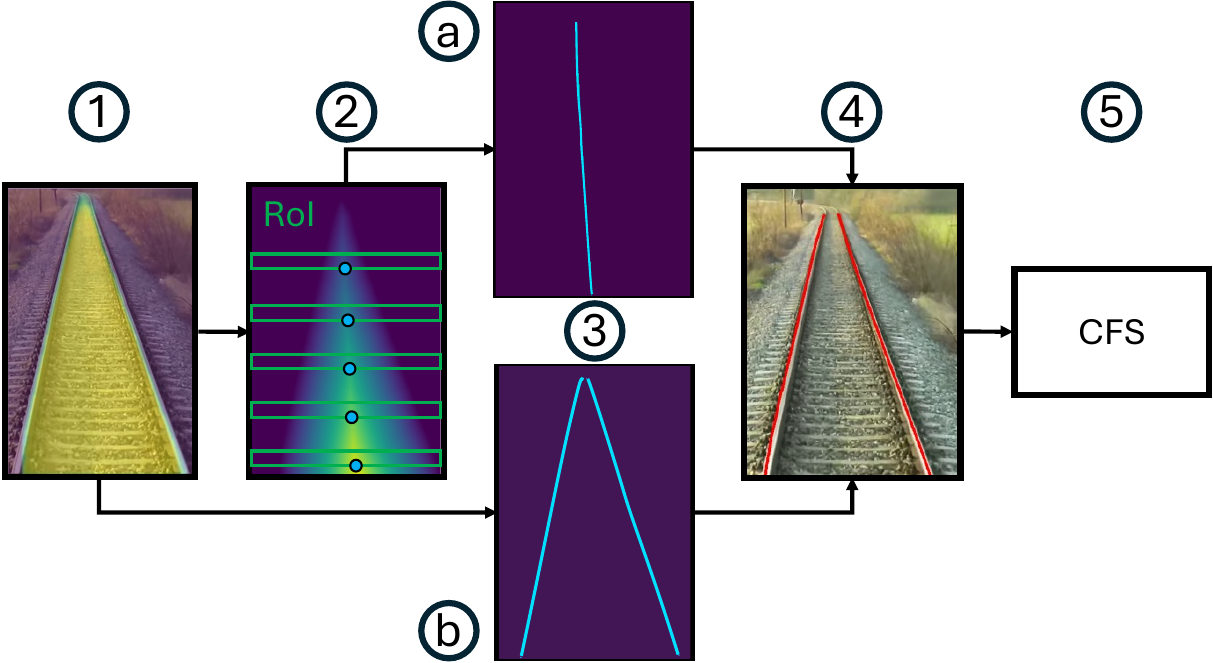}
            \caption[Schematic of the polyline-pair annotation algorithm for railway tracks with label "rail".]{Schematic of the polyline-pair annotation algorithm for railway tracks with label "rail". All images have been cropped to fit only one rail track instance exemplarily. For the sake of clarity, the first step of the pipeline shows both the input image to \ac{msrcnn} and the predicted instance segmentation mask overlayed on top of each other.}
            \label{Grafik: Polyline-pair annotation pipeline}
        \end{figure}

        First, \ac{msrcnn} predicts an instance segmentation mask for each detected rail object.
        We convert the soft mask into a binary mask using Otsu thresholding \cite{otsu1979threshold} and \ac{roi}-based histogram analysis.
        Spurious regions within the \ac{roi} are removed using a blob artifact filtering step.

        Second, we apply a distance transform \cite{strutz_distance_2023} to the binary mask to obtain a distance map.
        The rail centerline is extracted by sliding a window along the \ac{roi} and selecting local maxima.
        Specifically, for each image row, we choose the pixel with the highest pixel value, representing the center of the rail track at that row.
        Using this approach, we obtain the centerline coordinates as a set of $(x, y)$ points, regardless of track curvature.

        Third, we fit the extracted centerline coordinates with a fourth-degree polynomial to smooth outliers without excessive oscillation.
        The fitted points are then sorted by ascending $y$-coordinate to enable sequential processing (3a).
        In parallel, we extract and aggregate contour $(x, y)$ coordinates from the binary mask to represent both rail boundaries (3b).

\begin{figure}[!h]
            \centering
            \includegraphics[width = 0.36\imgWidth, keepaspectratio = true]{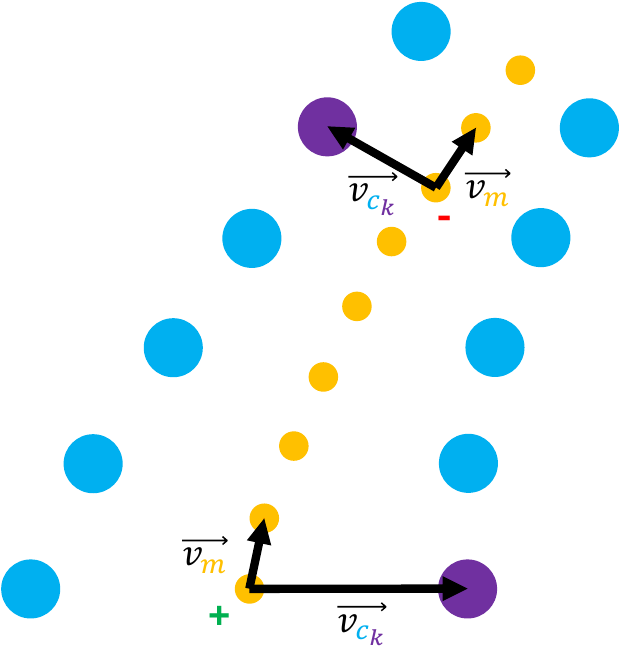}
            \caption[Functionality behind the cross-product point side membership decisioning algorithm.]{Functionality behind the cross-product point side membership decisioning algorithm. The contour coordinates are represented as large light-blue dots, whereas the middle line coordinates are represented by small yellow dots. The contour points for which the side membership has to be exemplarily decided are shown as large violet dots.}
            \label{Grafik: Side decisioning algorithm}
        \end{figure}

        Because a RailSem19 polyline-pair consists of two polylines \cite{zendel2019railsem19}, contour points must be assigned to one track side.
        We therefore classify each contour coordinate relative to the fitted centerline from step 3a.
        The assignment is performed using a 2D cross-product side-membership test, illustrated in Figure \ref{Grafik: Side decisioning algorithm}.

        For each contour point $c_k=(x_{c_k},y_{c_k})$, we select the nearest centerline point $m$ as reference.
        We define $\overrightarrow{v_{c_k}}$ from $m$ to $c_k$ and $\overrightarrow{v_m}$ from $m$ to its successor.
        A preliminary indicator $s'_{c_k}$ is computed as the cross product of the L2-normalized vectors:

        \begin{equation}
            s'_{c_k} = \frac{\overrightarrow{v_m}}{\left\lVert \overrightarrow{v_m} \right\rVert _2} \times \frac{\overrightarrow{v_{c_k}}}{\left\lVert \overrightarrow{v_{c_k}} \right\rVert _2}
            \label{equation: Side decisioning preliminary indicator formula}
        \end{equation}

        The side membership is then obtained from the sign of $s'_{c_k}$:

        \begin{equation}
            s_{c_k} =  \text{sign}(s'_{c_k}) = \begin{cases}
                1: \text{right} & \text{if } s'_{c_k} > 0 \\
                -1: \text{left} & \text{if } s'_{c_k} < 0 \\
                0: \text{error} & \text{if } s'_{c_k} = 0
            \end{cases}                    
            \label{equation: Side decisioning formula}
        \end{equation}

        Points with $s_{c_k}=1$ are assigned to the right polyline, and $s_{c_k}=-1$ to the left polyline.
        If $s_{c_k}=0$, the point is treated as invalid and excluded from both polylines.
        We apply this procedure to all contour points to construct the complete RailSem19 polyline-pair annotation.
        The resulting annotation is exemplarily shown in Figure \ref{Grafik: Polyline-pair annotation pipeline} with both polylines drawn as red lines on top of the input image.
        In the last step of the annotation module (step 5 in Figure \ref{Grafik: Polyline-pair annotation pipeline}), the generated annotations are filtered by the network quality indicators and or our own \ac{cfs}-score.

        The \ac{cfs}-score is a custom quality indicator for automatically generated rail-track polyline-pair annotations.
        It adapts the $F_\beta$ score \cite{peter_review_2023} to assess geometric consistency using contour agreement.
        The score is defined as:

        \begin{equation}
        F_\beta = \frac{(1 + \beta^2) \cdot P \cdot R}{\beta^2 \cdot P + R}
        \label{equation: F-Beta formula}
        \end{equation}

        where $F_\beta$ is the weighted harmonic mean of precision (P) and recall (R) \cite{peter_review_2023}.
        Precision and recall are computed from pointwise matches between two contour sets:

        \begin{equation}
        P = \frac{TP}{TP + FP} \quad \text{and} \quad R = \frac{TP}{TP + FN}
        \label{equation: Precision and Recall formula}
        \end{equation}

        Here, TP counts contour points consistent with the reference contour, FP counts erroneous assignments, and FN counts missing rail parts.
        We obtain these counts by overlaying the predicted rail contour prior to conversion with the contour reconstructed after the annotation pipeline.
        False positives typically arise from side-membership or centerline extraction errors, whereas false negatives reflect incomplete coverage.
        We set $\beta = 0.75$ empirically to prioritize precision and penalize incorrect rail assignments.

\begin{figure}[!h]
            \centering
            \includegraphics[width = 0.43\imgWidth, keepaspectratio = true]{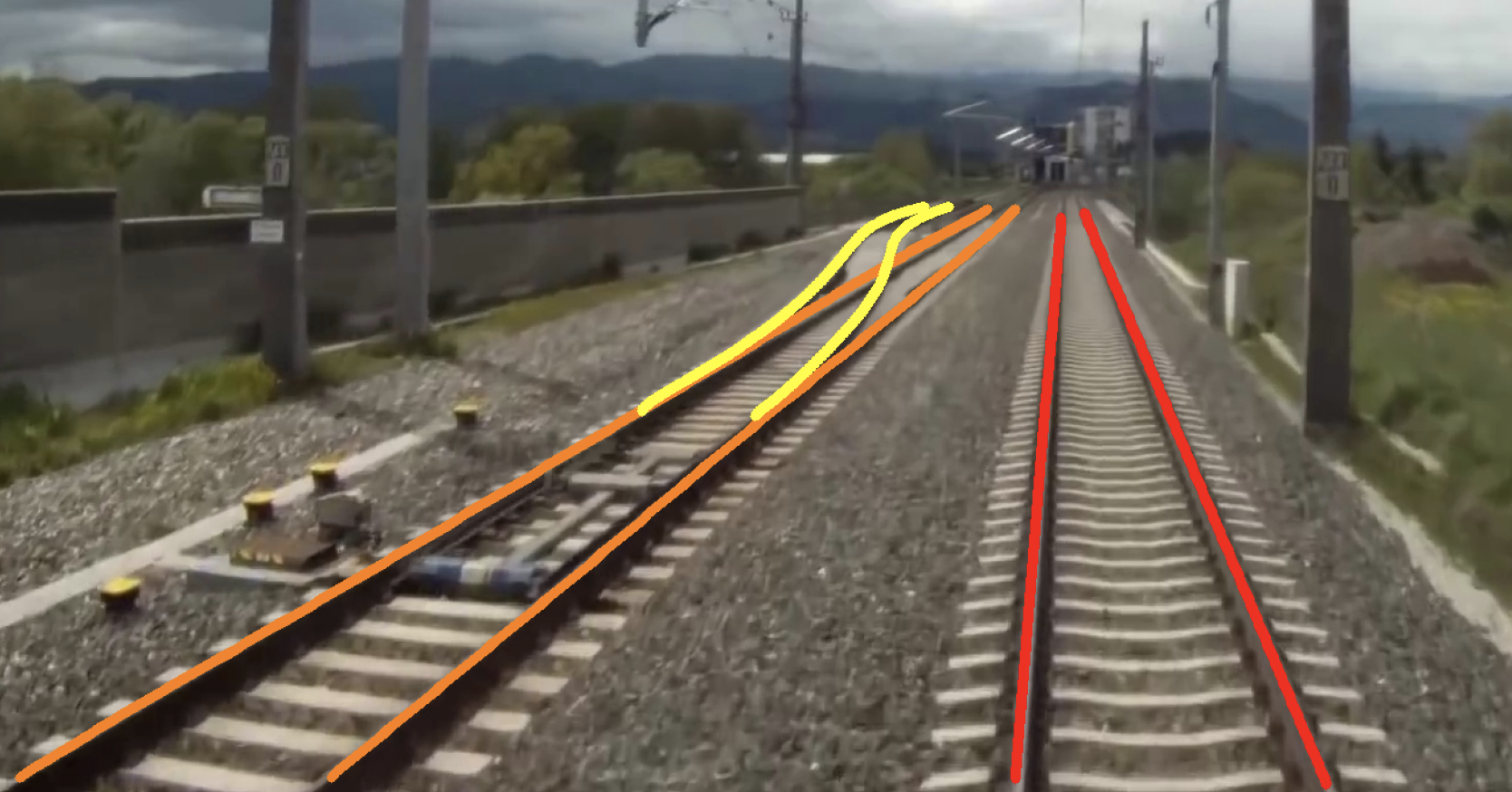}
            \caption{Qualitative ARIDA output on "rs00057.jpg" from the RailSem19 dataset \cite{zendel2019railsem19}, shown here only for the polyline-pair rail annotation type as a showcase. The source image was cropped to emphasize the central content. Native ARIDA annotation information markers were deactivated. The maskIoU scores for the detected three rail instances are 0.98, 0.95 and 0.93, while the CFS scores are 0.96, 0.94 and 0.97 respectively (in colored order of red, orange and yellow), also indicating the observed high annotation quality.}
            \label{Grafik: Polyline-pair annotation example}
        \end{figure}

        \ac{cfs} serves as a no-ground-truth quality gate that rejects low-quality polyline-pair annotations using a threshold.
        Combined with cls and mask\ac{iou} filtering, it enables largely autonomous generation of high-quality railway annotations.
        While Figures \ref{Grafik: Polyline-pair annotation pipeline} and \ref{Grafik: Side decisioning algorithm} illustrate the conversion procedure schematically, Figure \ref{Grafik: Polyline-pair annotation example} presents a qualitative output example on a real, automatically annotated RailSem19 image, showing per-instance rail splitting and the curvature-following capability of the proposed polyline-pair generation algorithm.

    \subsection{\acl{al} and \acl{tbal}}
        \ac{al} is an iterative training paradigm that reduces labeling effort by selecting informative samples for annotation.
        The underlying hypothesis is that a model reaches a target performance using fewer labels when it queries the most relevant unlabeled examples.
        To operationalize this idea, \ac{al} alternates between training the model and requesting new annotations from a human oracle. \cite{settles2009active}\cite{li2025deepactivelearning}

        To enable systematic querying, the available dataset $\mathcal{D}$ is partitioned into a labeled subset $\mathcal{D_\mathcal{L}}$ and an unlabeled subset $\mathcal{D_\mathcal{U}}$.
        The model is initialized on $\mathcal{D_\mathcal{L}}$ and then applied to candidate images sampled from $\mathcal{D_\mathcal{U}}$.
        A query strategy is required to assign each candidate a relevancy score and to select the next images for manual annotation.
        After annotation, the queried samples are added to $\mathcal{D_\mathcal{L}}$ and the model is retrained, forming an \ac{al}-cycle. \cite{settles2009active}\cite{li2025deepactivelearning}
        Within \ac{arida}, this cyclic optimization is essential to maintain reliable automatic annotations as image difficulty increases.
        Query strategies are commonly grouped into representativeness-based and informativeness-based methods \cite{settles2009active}\cite{li2025deepactivelearning}.

        Representativeness strategies seek diverse samples using clustering or density estimation \cite{settles2009active}\cite{li2025deepactivelearning}, which is impractical for high-dimensional images.
        We therefore use informativeness-based querying, which selects samples where the model is uncertain.
        However, standard uncertainty measures such as class-probability entropy \cite{li2025deepactivelearning} are not directly suited for two-stage instance segmentation.
        \ac{msrcnn} outputs a variable number of instances and distinct confidence indicators for classification and mask quality.
        Consequently, a single probability-based uncertainty score cannot capture stage disagreement or low mask fidelity. \cite{xu_active_2022}

        We therefore adopt the \ac{tbal} strategy by Xu et al. \cite{xu_active_2022} and specifically tailor it to \ac{msrcnn}, by omitting the box-quality term, which is not needed for the instance segmentation-focused annotation task in \ac{arida}.
        For each image $m$ and detected instance $o$, we form the indicator tuple \cite{xu_active_2022}:

        \begin{equation}
            T^o_m = \langle cls^o_m, mask^o_m \rangle
            \label{equation: Indicators tuple}
        \end{equation}

        where $cls^o_m$ denotes the classification score and $mask^o_m$ the predicted mask\ac{iou} score.
        These indicators quantify semantic confidence and estimated mask quality per object instance respectively. \cite{xu_active_2022}
        Instance informativeness is then computed from their mean and standard deviation \cite{xu_active_2022}:

        \begin{equation}
            I^o_m = 1 - \mean(T^o_m) + \std(T^o_m)
            \label{equation: Informativeness object}
        \end{equation}

        where low average confidence and high indicator disagreement increase $I^o_m$ \cite{xu_active_2022}.
        Image informativeness is then obtained by aggregating over all instances \cite{xu_active_2022}:

        \begin{equation}
            I_m = \sum_{o} I^o_m
            \label{equation: Informativeness image}
        \end{equation}

        This definition prioritizes images containing many uncertain instances and thus high expected labeling benefit.
        If no instances are detected, we propose to heuristically set $I_m = 1000.0$ to prioritize likely failure cases.
        This value is chosen empirically to exceed typical informativeness scores, obtainable from a maximum of 100 detectable objects using the general \ac{msrcnn} architecture \cite{he_mask_2018}.
        The highest-ranked images are sent to the oracle for annotation and incorporated into $\mathcal{D_\mathcal{L}}$.
        The procedure repeats until a predefined stopping criterion, such as an annotation budget, is reached. \cite{xu_active_2022}\cite{settles2009active}
        In case of \ac{arida}, this budget is dynamically controlled by the sorting module through the \ac{qqp} mechanism described in the next Section.

    \subsection{Sorting module and \acl{qqp}}
        \ac{al} in \ac{arida} improves \ac{msrcnn} cyclically by querying only high-value samples for manual annotation.
        This mechanism is particularly important for images that cannot be annotated automatically under set quality constraints.
        To further reduce effort, oracle queries should be issued only when automatic annotation is insufficient due to missing model capability.
        \ac{arida} therefore separates "easy" and "hard" images within each annotation cycle.

        All images are first processed by the automatic annotation module using fixed acceptance criteria.
        Images that satisfy these criteria are removed from the oracle queue.
        Unsuccessful cases remain candidates for manual annotation and are prioritized by informativeness.

        The sorting module receives input images and corresponding \ac{msrcnn} predictions (s. Figure \ref{Grafik: Framework Architecture}).
        It computes the \ac{tbal} informativeness score and ranks images accordingly.
        Low-informativeness images are processed first for automatic annotation, whereas high-informativeness images are reserved for the oracle.
        The number of successful automatic annotations in a cycle is denoted by $K_A$.

        The oracle query budget $K_{Or}$ is adapted using \ac{qqp}.
        It decreases monotonically with increasing $K_A$ according to our proposed formula:

        \begin{equation}
            K_{Or}(K_A)=\Bigl\lfloor
            K_{Or_{max}}-\frac{K_A}{K_{A_{max}}}\bigl(K_{Or_{max}}-K_{Or_{min}}\bigr)
            \Bigr\rfloor
        \label{equation: Query quantity parameterization}
        \end{equation}

        Here, $K_{Or_{min}}$ and $K_{Or_{max}}$ define lower and upper bounds for oracle queries.
        Parameter $K_{A_{max}}$ specifies the maximum number of automatic annotations per cycle.
        When $K_A$ is small, $K_{Or}$ increases to provide additional labeled data for difficult cases.
        When $K_A$ is large, $K_{Or}$ is reduced because little oracle intervention is required, thus current model knowledge suffices.

%% file: Kapitel/Results/Results.tex
\section{Results}
    \label{section: Results}

    This section presents the results of evaluating the \ac{arida}-framework on the disjoint RailSem19 and sequential \ac{osdar23} datasets.
    Also, the contribution of the \ac{tbal}-based query strategy to the overall performance and effort reduction is analyzed in a dedicated evaluation step.

    In the first part (A) of this section, the overall parameterization and the individual evaluation steps of the \ac{arida}-framework are explained. 
    Subsequently, the results for each evaluation step are presented in the second part (B).

    \subsection{Parameterization and evaluation procedure}
        \label{subsection: parameterization and evaluation procedure}

        We evaluate \ac{arida} under two complementary railway data regimes.
        First, we study disjoint and heterogeneous imagery, where viewpoints and capture settings vary strongly.
        This setting stresses model generalization and makes sample selection harder.
        We use RailSem19 to represent this regime.

        Second, we study sequential and highly redundant imagery from a fixed capture setup.
        This setting reflects practical data acquisition along a route and highlights the risk of over-annotation.
        We use \ac{osdar23} to represent this regime, though all scenery will be used for both training and evaluation.
        Based on these regimes, we perform three evaluation steps.

        \begin{enumerate}
            \item \ac{arida} on disjoint RailSem19 data to test robustness.
            \item A \ac{tbal} study on RailSem19 to isolate query ranking effects.
            \item \ac{arida} on sequential \ac{osdar23} data to test redundancy handling.
        \end{enumerate}

        All experiments train \ac{msrcnn} on a single-GPU server with limited memory.
        We therefore fix the per-iteration batch size to 1 and apply gradient accumulation to obtain an effective batch size of 2.
        This choice keeps training stable without substantially increasing operational time.
        We use AdamW with a learning rate of $10^{-4}$ throughout.
        The complete training configuration is summarized in Table~\ref{tab:train_params}.

        \begin{table}[htbp]
            \centering
            \caption{General training parameters (constant across experiments).}
            \label{tab:train_params}
            \begin{tabular}{p{4.3cm} p{3.5cm}}
                \hline
                \textbf{Parameter} & \textbf{Setting} \\
                \hline
                Optimizer & AdamW \\
                Learning rate & $10^{-4}$ \\
                Batch size & 1 \\
                Gradient accumulation & 2 (effective batch size) \\
                Pretraining epochs & 500 \\
                Retraining epochs per \ac{al} cycle & 100 \\
                \hline
            \end{tabular}
        \end{table}

        We deliberately evaluate the hardest track-annotation setting by restricting the task to one class and one primitive.
        Concretely, we only use polyline-pair annotations for training and for \ac{arida}'s annotation actions.
        To ensure that the learning signal is rail-relevant, we filter samples by their polyline-pair label "rail" for RailSem19 and "track" for \ac{osdar23}.
        For each run, we construct different labeled pool $\mathcal{D}_\mathcal{L}$ sizes, the unlabeled pool $\mathcal{D}_\mathcal{U}$ and the evaluation set $\mathcal{D}_\mathcal{E}$ by random sampling with seed 1234 without overlap.
        This disjoint construction avoids leakage that would underestimate the oracle workload.
        Pool sizes, dataset-specific adjustments, and the fixed sampling seed are listed in Table~\ref{tab:al_params}.

        \begin{table}[htbp]
            \centering
            \caption{\ac{al} parameterization and pool construction.}
            \label{tab:al_params}
            \begin{tabular}{p{4.3cm} p{3.5cm}}
                \hline
                \textbf{Parameter} & \textbf{Setting} \\
                \hline
                Default unlabeled pool size & $|\mathcal{D}_\mathcal{U}|=1000$ \\
                \\
                RailSem19 evaluation set size & $|\mathcal{D}_\mathcal{E}|=500$ \\
                \\
                \ac{osdar23} evaluation set size & $|\mathcal{D}_\mathcal{E}|=200$ \\
                \ac{osdar23} largest-run unlabeled pool & $|\mathcal{D}_\mathcal{U}|=748$ (remaining data) \\
                \hline
            \end{tabular}
        \end{table}

        For both RailSem19 and \ac{osdar23}, we fix the unlabeled pool size to 1000 images.
        As \ac{osdar23} contains fewer total images, the evaluation set is reduced to 200 images in comparison to RailSem19's 500 images.
        This allows for a sufficiently large unlabeled pool even when the initial labeled set is large.

        Each evaluation step begins with a pretraining phase on $\mathcal{D}_\mathcal{L}$ for 500 epochs.
        After every \ac{al}-cycle, the model is retrained for 100 epochs, since only a small number of images is added per cycle.
        Performance is measured after each cycle on $\mathcal{D}_\mathcal{E}$ using torchmetrics segmentation \ac{map} \cite{noauthor_mean_nodate} with its standard \ac{iou} sweep from $0.5$ to $0.95$ \cite{noauthor_mean_nodate}.
        This yields learning curves as a function of the labeled set size $N(\mathcal{D}_\mathcal{L})$.

        \ac{arida}'s annotation module accepts automatic annotations only if their quality is sufficiently high.
        We gate predictions using three indicators: the classification confidence, the predicted mask\ac{iou} score, and the \ac{cfs} score.
        The thresholds (Table~\ref{tab:anno_params}) were set empirically to reject low-quality automatic labels early and to route uncertain cases to the oracle instead.
        This trades some automation for higher label reliability and cleaner feedback during \ac{al}.

        \begin{table}[htbp]
            \centering
            \caption{Annotation module acceptance thresholds.}
            \label{tab:anno_params}
            \begin{tabular}{p{4.3cm} p{3.5cm}}
                \hline
                \textbf{Parameter} & \textbf{Setting} \\
                \hline
                Classification score threshold & $\ge 0.99$ \\
                Predicted mask\ac{iou} threshold & $\ge 0.90$ \\
                \ac{cfs} threshold & $\ge 0.70$ \\
                \hline
            \end{tabular}
        \end{table}

        Oracle interaction is controlled by \ac{qqp}, which bounds manual intervention per cycle while still allowing correction when needed.
        In each cycle, \ac{qqp} limits the number of oracle-provided annotations to the interval $[K_{Or_{min}}, K_{Or_{max}}]$ and caps the number of automatically generated annotations.
        We simulate the oracle by revealing hidden ground truth labels from $\mathcal{D}_\mathcal{U}$, because the goal is to isolate the contribution of \ac{arida} components without human variability.
        The \ac{qqp} settings used in all experiments are given in Table~\ref{tab:qqp_params}.

        \begin{table}[htbp]
            \centering
            \caption{\ac{qqp} settings for oracle querying.}
            \label{tab:qqp_params}
            \begin{tabular}{p{4.3cm} p{3.5cm}}
                \hline
                \textbf{Parameter} & \textbf{Setting} \\
                \hline
                Oracle-query lower bound & $K_{Or_{min}}=0$ \\
                Oracle-query upper bound & $K_{Or_{max}}=10$ \\
                \\
                Max. automatic annotations per cycle & $K_{A_{max}}=100$ \\
                \hline
            \end{tabular}
        \end{table}

        We run three evaluation steps to validate \ac{arida} and its components.

        \begin{enumerate}
            \item \textbf{Disjoint RailSem19:} $N_{\text{initial}}(\mathcal{D}_\mathcal{L}) \in \{100, 200, 500\}$ compared to supervised baselines trained on $\{100, 200, 500\}$ labeled images.
            \item \textbf{\ac{tbal} study:} RailSem19 is used again, comparing \ac{tbal}-high, random, and \ac{tbal}-low score selection under the best-performing configuration from step (1).
            \item \textbf{Sequential \ac{osdar23}:} $N_{\text{initial}}(\mathcal{D}_\mathcal{L}) \in \{10, 20, 50, 100, 200, 500\}$ compared to supervised baselines trained on $\{20, 50, 100, 200, 500, 1000\}$ labeled images.
        \end{enumerate}

        For steps (1) and (3), we consider \ac{al} beneficial if the \ac{al} learning curve reaches a supervised baseline \ac{map} while using fewer labeled images.
        For step (2), we consider \ac{tbal} effective if the highest-\ac{tbal} strategy yields higher \ac{map} at a given labeled-set size than random or lowest-\ac{tbal} selection.
        In addition to accuracy, we quantify manual workload reduction by tracking the total number of oracle annotations $N(K_{Or})$.
        We compute the aggregate effort reduction relative to the unlabeled pool size as:

        \begin{equation}
        \eta_{\text{reduction}} = 1 - \frac{N_{\text{initial}}(\mathcal{D}_L) + N(K_{Or})}{N(\mathcal{D}_U)}
        \end{equation}

        This measure accounts for both the initial bootstrapping effort and subsequent oracle interventions across cycles.

    \subsection{Results on effort reduction}
        \subsubsection{Evaluation of \ac{arida} on disjoint RailSem19 data}
\begin{figure}[htbp]
                \centering
                \includegraphics[width=1.0\linewidth, keepaspectratio=true]{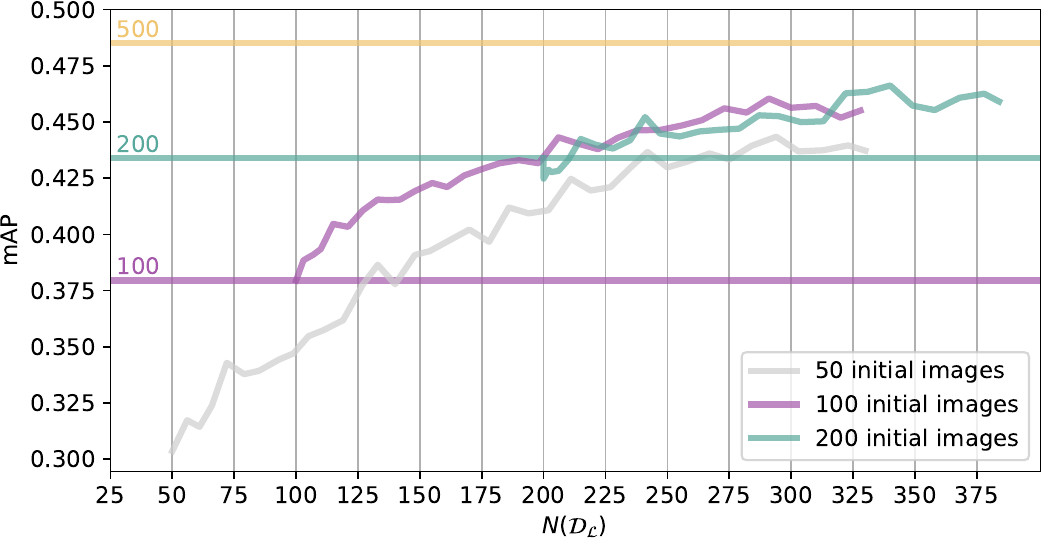}
                \caption{Segmentation \ac{map} progression on disjoint RailSem19 data for three initial label budgets (50, 100 and 200). The x-axis shows the current labeled-set size \(N(\mathcal{D}_L)\). The horizontal markers indicate supervised baselines trained on 100, 200, and 500 labeled images.}
                \label{fig:rs19_map}
            \end{figure}

            This experiment evaluates \ac{arida} on disjoint RailSem19 imagery to address \textbf{RQ} on reducing manual annotation effort while maintaining model utility. 
            Also, both hypotheses regarding overall effort reduction through \ac{arida} (hypothesis 1) and \ac{al} reaching supervised accuracy with fewer manual labels (hypothesis 2) are tested in this experiment.
            Figure~\ref{fig:rs19_map} reports the segmentation \ac{map} progression over the current labeled-set size.

            Across all three runs, \ac{map} increases as additional oracle-labeled samples enter $\mathcal{D}_\mathcal{L}$.
            The run initialized with 50 labeled images shows the steepest improvement, but converges at the lowest final accuracy ($\ac{map} \approx \perc{43.7}$).
            Initializing with 100 or 200 labeled images yields higher \ac{map} throughout most of the process.
            Notably, the 100-initial configuration reaches comparable performance to the 200-initial run after roughly 200 – 315 labeled images.
            This behavior reduces over-annotation risk by avoiding a large up-front labeling commitment for small gains.

            Comparing the \ac{al} curves to supervised baselines trained on fixed budgets (100, 200, and 500 images), the 50-initial run does not exceed any supervised baseline while utilizing less labeled data. 
            Thus, hypothesis 2 is not supported for this scarce initialization.
            In contrast, the 100-initial run matches the supervised 200-image baseline at exactly 200 images and also surpasses the 200-image baseline mostly thereafter, confirming hypothesis 2 under moderate prior knowledge.
            None of the runs reach the 500-image baseline because the available unlabeled pool is exhausted before that budget.
            Importantly, for the early plateauing 50- and 100-initial runs, stopping after 325 labeled images in comparison to 390 labeled images for the 200-initial run results in a modest 5 percentage point drop in \ac{map}, while saving 65 manual annotations.
            This illustrates how \ac{al} can act as a safeguard against over-annotation when returns diminish.

            Table~\ref{tab:rs19_effort} summarizes the resulting effort reduction per run.

            \begin{table}[htbp]
                \caption{Manual and automatic annotation counts, and resulting effort reduction, for disjoint RailSem19 runs.}
                \label{tab:rs19_effort}
                \centering
                \begin{tabular}{c<{\hspace{0.3cm}} c<{\hspace{0.3cm}} c<{\hspace{0.3cm}} c}
                    \toprule
                    $N_{\text{initial}}(\mathcal{D}_L)$ & $N(K_{Or})$ & $N(K_{A})$ & $\eta_{\text{reduction}}$ \\
                    \midrule
                    50  & 280 & 720 & 67.0\% \\
                    100 & 228 & 772 & \textbf{67.2\%} \\
                    200 & 184 & 816 & 61.6\% \\
                    \bottomrule
                \end{tabular}
            \end{table}

            All configurations reduce manual effort substantially, with $\eta_{\text{reduction}}$ between \perc{61.6} and \perc{67.2}.
            The 100-initial run achieves the highest effort reduction (\perc{67.2}) while also providing the most favorable accuracy–cost trade-off.
            Starting from 200 labeled images reduces additional oracle queries, but lowers $\eta_{\text{reduction}}$ due to the doubled initial manual investment.
            In disjoint, diverse scenes, this pattern highlights the practical tension between small accuracy improvements and unnecessary up-front labeling.

            On disjoint RailSem19 data, \ac{arida} consistently reduces manual labeling by about two thirds.
            The \ac{al} sample-efficiency hypothesis is supported for the 100-initial configuration, but not for 50 initial labels.
            Overall, 100 initial labels best balances accuracy, effort reduction, and over-annotation risk, thus making it the optimal parameterization for subsequent evaluations and practical use on disjoint data.

        \subsubsection{Evaluation of the \acs{tbal}-component}
            This experiment isolates the \ac{tbal}-based query strategy within \ac{arida} to assess its contribution to the paper’s \textbf{RQ}.
            In line with hypothesis 1, we measure how \ac{tbal} affects overall manual effort through $N(K_{Or})$ and $\eta_{\text{reduction}}$.
            In line with hypothesis 2, we analyze whether \ac{tbal} improves sample efficiency by reaching a given \ac{map} with fewer labeled images.
            Because RailSem19 is highly diverse, avoiding over-annotation is especially relevant once gains plateau.

\begin{figure}[htbp]
                \centering
                \includegraphics[width=1.0\linewidth, keepaspectratio=true]{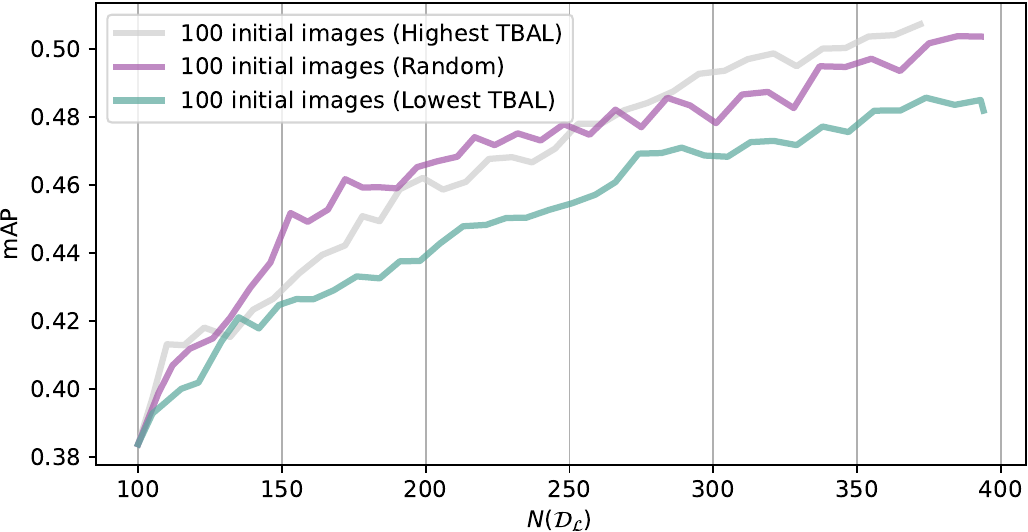}
                \caption{Segmentation \ac{map} progression on disjoint RailSem19 data with fixed $N_{\text{initial}}(\mathcal{D}_L) = 100$ and three \ac{al} sampling strategies. The x-axis shows the current labeled-set size $N(\mathcal{D}_L)$.}
                \label{fig:tbal_rs19_map}
            \end{figure}

            We reuse the best-performing disjoint-data configuration from Evaluation 1 with $N_{\text{initial}}(DL) = 100$ and compare three sampling strategies: highest \ac{tbal} score (most informative), random, and lowest \ac{tbal} score (least informative).
            Figure~\ref{fig:tbal_rs19_map} shows the segmentation \ac{map} progression over the current labeled-set size $N(\mathcal{D}_\mathcal{L})$.

            In the early phase (about 100 – 125 labeled images), selecting highest-\ac{tbal} samples yields a small \ac{map} advantage over lowest-\ac{tbal} selection (up to 2 percentage points). 
            Compared to random querying, the difference is minor, but the highest-\ac{tbal} curve tends to improve more steadily.
            Around $N(\mathcal{D}_\mathcal{L}) \approx 125$, all three curves meet near $\ac{map} \approx 0.42$, indicating that query choice has limited influence at this stage.
            As labeling proceeds, lowest-\ac{tbal} selection shows the weakest growth, which translates into lower accuracy gain per additional annotation.

            Random querying briefly outperforms highest-\ac{tbal} by up to 2 percentage points near $N(\mathcal{D}_\mathcal{L}) \approx 150$, but this advantage does not persist. 
            Beyond roughly $N(\mathcal{D}_\mathcal{L}) \approx 250$, the highest-\ac{tbal} strategy becomes the most efficient again and ends with the best final \ac{map}.

            At the end of the process, highest-\ac{tbal} reaches approximately \perc{50} \ac{map}, compared to \perc{49} for random and \perc{48} for lowest-\ac{tbal}.
            This difference is small in absolute terms, yet it is practically relevant when annotation should stop at a target quality to avoid over-annotation.
            For example, to reach \perc{49} \ac{map}, highest-\ac{tbal} needs about 290 labeled images, whereas random querying requires about 330.
            In other words, informativeness-guided selection can save roughly 40 manual labels at a fixed target performance, directly supporting the goal of reducing unnecessary annotation work.
            Table~\ref{tab:tbal_effort} confirms that this performance advantage coincides with higher end-to-end efficiency.

            \begin{table}[htbp]
                \centering
                \caption{Oracle and automatic annotation counts, and resulting effort reduction, for the \ac{tbal} comparison on RailSem19.}
                \label{tab:tbal_effort}
                \begin{tabular}{l c c c}
                    \toprule
                    Sampling strategy & $N(K_{Or})$ & $N(K_A)$ & $\eta_{\text{reduction}}$ \\
                    \midrule
                    Highest \ac{tbal} & 272 & 728 & \textbf{62.8\%} \\
                    Random & 293 & 707 & 60.7\% \\
                    Lowest \ac{tbal} & 294 & 706 & 60.6\% \\
                    \bottomrule
                \end{tabular}
            \end{table}

            With the same initial pool of 100 labeled images, highest-\ac{tbal} queries the oracle least ($N(K_{Or}) = 272$) and enables the most automatic annotations ($N(K_{A}) = 728$), yielding the best effort reduction ($\eta_{\text{reduction}} = \perc{62.8}$).
            Random querying and lowest-\ac{tbal} selection require about 21 – 22 more oracle annotations, reducing efficiency to $\eta_{\text{reduction}} \approx \perc{60.7}$. 
            These results align with hypothesis 1 by demonstrating measurable workload differences attributable to the query strategy.
            They also support hypothesis 2 in the sense that highest-\ac{tbal} reaches target \ac{map} values with fewer labeled images than random selection.

        \subsubsection{Evaluation of \ac{arida} on sequential \ac{osdar23} data}
            This experiment evaluates \ac{arida} on time-sequential, potentially redundant railway imagery from \ac{osdar23}.
            In this regime, many frames share similar content, so over-annotation is a primary practical risk.

\begin{figure}[htbp]
                \centering
                \includegraphics[width=1.0\linewidth, keepaspectratio=true]{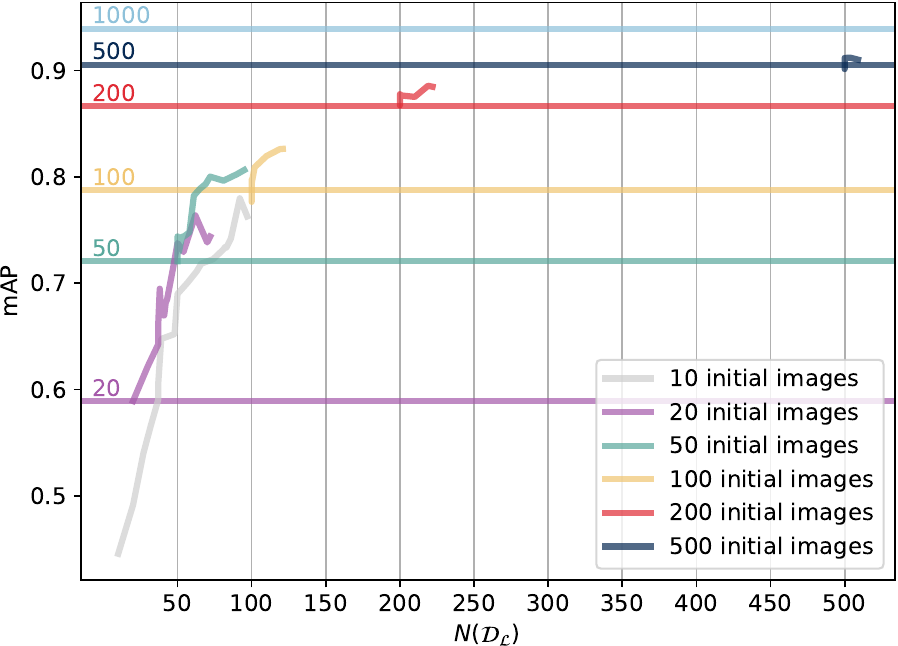}
                \caption{Segmentation \ac{map} progression on sequential \ac{osdar23} data for different initial label budgets. The x-axis shows the current labeled-set size $N(\mathcal{D}_L)$. Horizontal markers indicate supervised baselines trained on fixed budgets (20 - 1000 images).}
                \label{fig:osdar23_map}
            \end{figure}

            Figure~\ref{fig:osdar23_map} shows segmentation \ac{map} over the current labeled-set size $N(\mathcal{D}_\mathcal{L})$ for initial budgets from 10 to 500 images.

            The key pattern is that small initial budgets learn fastest, because early \ac{al} cycles still expose new information despite redundancy. 
            In particular, the 50-initial run reaches about \perc{79} \ac{map} (the supervised 100-label level) after roughly 60 labeled images, confirming hypothesis 2 at a practically relevant operating point and avoiding about 40 potentially redundant labels.
            The 20-initial run tracks the 50-initial curve closely, offering a safer starting point when over-annotation must be minimized. 
            By contrast, 10 initial labels can be the least risky upfront investment, but it reaches the supervised 20-label baseline only after roughly 40 labels and therefore does not satisfy hypothesis 2 early on.

            For larger initial pools, redundancy dominates and marginal gains per new label shrink.
            The 100-initial run does not reach the 200-label baseline, yet \ac{al} still prevents unnecessary labeling by stopping near 125 images, saving about 75 labels for only a small accuracy gain.
            The same effect strengthens at 200 and 500 initial labels, where runs fail to catch the next larger baselines and mainly demonstrate \ac{al} as a safeguard against over-annotation. 
            This behavior matches the intended role of \ac{al} in sequential streams, where performance plateaus well before the remaining frames are exhausted.

            Table~\ref{tab:osdar23_effort} supports hypothesis 1 by showing very large effort reductions for small initial budgets.

            \begin{table}[htbp]
                \centering
                \caption{Oracle and automatic annotation counts, and resulting effort reduction, for sequential \ac{osdar23} runs.}
                \label{tab:osdar23_effort}
                \begin{tabular}{c c c c}
                    \toprule
                    $N_{\text{initial}}(\mathcal{D}_L)$ & $N(K_{Or})$ & $N(K_A)$ & $\eta_{\text{reduction}}$ \\
                    \midrule
                    10  & 87 & 913 & 90.3\% \\
                    20  & 52 & 948 & \textbf{92.8\%} \\
                    50  & 45 & 955 & 90.5\% \\
                    100 & 21 & 979 & 87.9\% \\
                    200 & 22 & 978 & 77.8\% \\
                    500 & 9  & 239 & 31.9\% \\
                    \bottomrule
                \end{tabular}
            \end{table}

            The best overall reduction occurs at 20 initial labels with $\eta_{\text{reduction}} = \perc{92.8}$ while 10 and 50 initial labels remain above \perc{90}. 
            Efficiency deteriorates as the initial pool grows, dropping to \perc{77.8} at 200 and to \perc{31.9} at 500 initial labels, because upfront labeling dominates the budget.

            For time-sequential railway data, we recommend starting with 20 – 50 labeled images.
            The 20-initial setting is a strong default when over-annotation is a priority, while 50-initial provides a robust "fast ramp" to high \ac{map}.
            It is recommended to avoid large initial pools with 200 or more labeled images on sequential streams, because redundancy limits returns and increases over-annotation risk. 

            To conclude, on \ac{osdar23} \ac{arida} achieves up to \perc{92.8} effort reduction, supporting hypothesis 1.
            Hypothesis 2 is best supported for 20 – 50 initial labels, where \ac{al} reaches strong accuracy with fewer manual annotations.
            Over-annotation is mitigated most effectively by conservative initial budgets and \ac{al}-based early stopping.

%% file: Kapitel/Discussion/Discussion.tex
\section{Discussion}
    \label{section: Discussion}

    Modern railway perception requires large, high-quality datasets, yet manual annotation remains costly and operationally constrained.
    To address this problem, \ac{arida} combines automatic annotation, domain-specific quality gating, and \ac{al} so that human effort is concentrated on uncertain or informative samples.
    Across all experiments, \ac{arida} substantially reduced manual effort, but its benefit depended strongly on dataset structure.
    The results suggest that \ac{arida} is most effective either when the model must first cross a competence threshold before automatic acceptance and uncertainty-based querying become reliable, or when sequential redundancy makes additional manual labeling increasingly inefficient.
    For disjoint and heterogeneous data, the main finding is the existence of such a competence threshold.

    In the present RailSem19 setting, an initial pool of about 100 labeled images provided the best trade-off between early model stability and upfront labeling cost.
    Below this level, strict acceptance gates reject many predictions and labeled-set growth depends mainly on oracle corrections, so uncertainty estimates are not yet sufficiently informative.
    Once this threshold is reached, both automatic acceptance and active selection become more effective.
    The \ac{tbal} study supports this interpretation: informativeness-based selection improves sample efficiency only after the detector has learned reasonably stable instance representations.
    For sequential and highly redundant data, the interpretation differs.

    Here, \ac{arida} is less important for discovering diverse supervision than for preventing over-annotation once the model has already captured most of the available information.
    On \ac{osdar23}, small initial pools of about 20 to 50 images reached strong performance quickly, while larger initial pools mainly increased upfront cost and reduced achievable effort savings.
    In this regime, \ac{al} acts primarily as a stopping and prioritization mechanism: easy samples are accepted early, and the remaining pool becomes progressively harder and less informative.
    The resulting performance plateau should therefore be interpreted as diminishing return rather than simple underperformance.

    Direct comparison with earlier related work remains limited because existing studies differ in evaluation setup, annotation workflow, and effort metric.
    This is especially true for the closest railway-specific upstream reference, Labels4Rails, which reports wall-clock per-image annotation-time reductions in a semi-automated workflow with continuous human supervision and manual amendment \cite{hiebert2025labels4rails}.
    By contrast, \ac{arida} targets reduction of manual intervention demand through quality-controlled automatic acceptance and \ac{al}, and measures effort through initial manual labels and oracle interventions relative to the unlabeled pool.
    Comparison with Labels4Rails is therefore informative at a qualitative level, but not directly quantitative.

    These conclusions should be interpreted within the limits of the present study.
    All experiments used a single random seed, the oracle was simulated from hidden ground truth rather than implemented with human annotators, only two datasets were evaluated, and the experiments were restricted to one class and one complex primitive.
    In addition, acceptance thresholds and \ac{qqp} settings were chosen empirically.
    The reported operating points should therefore be viewed as practically useful guidelines for this setup rather than universally optimal settings.
    Within these limits, the experiments suggest starting with a modest but nontrivial initial pool for disjoint data (roughly \perc{5} to \perc{10} of the unlabeled set) and a smaller initial pool for sequential data (about 20 to 50 images), while monitoring saturation and stopping when further gains become marginal.

%% file: Kapitel/Conclusion/Conclusion.tex
\section{Conclusion}
    \label{section: Conclusion}

    This paper answers the research question \textbf{RQ} on how to reduce manual effort while still producing high-quality, railway-specific multi-primitive image annotations.
    It introduces \ac{arida}, a semi-automated framework that converts \ac{msrcnn} instance predictions into RailSem-compatible primitives, including track polyline-pairs, and controls automatic track geometry annotation via the no-ground-truth \ac{cfs} acceptance gate.
    \ac{arida} further integrates an \ac{al}-loop with \ac{tbal}-based informativeness ranking and a \ac{qqp}, so that "easy" images are auto-accepted first and oracle effort is concentrated on genuinely hard, high-value samples.

    Empirically, \ac{arida} reduces manual labeling by roughly two thirds on disjoint RailSem19 data ($\eta_{\text{reduction}} \approx \perc{61.6}$ – \perc{67.2}), with 100 initial labels providing the best accuracy–over-effort trade-off.
    On sequential, redundant \ac{osdar23} data, \ac{arida} achieves up to \perc{92.8} effort reduction, and small initial pools (20 – 50 labels) learn fastest while mitigating over-annotation through early saturation and \ac{al}-driven stopping.
    For future work, some limitations should be addressed to test the generality of these conclusions.

    The most immediate next steps are multi-seed experiments, real human-in-the-loop evaluation with timing measurements, and extension to multi-class and multi-primitive workloads.
    It would also be useful to study the sensitivity of \ac{arida} to its acceptance thresholds and QQP settings more systematically, since these parameters directly shape both effort and accuracy.
    Finally, richer selection strategies, including temporal cues for sequential data, may further improve the balance between automatic annotation, oracle workload, and stopping decisions.